\documentclass[runningheads]{llncs}

\usepackage{eccv}

\usepackage{eccvabbrv}

\usepackage{graphicx}
\usepackage{booktabs}

\usepackage[accsupp]{axessibility}  %
\usepackage{amsmath}
\usepackage{subcaption}
\usepackage{makecell}%
\usepackage{multirow}
\usepackage{wrapfig}
\usepackage{bbm}

\usepackage[dvipsnames]{xcolor}

\DeclareMathOperator*{\argmin}{arg\,min}

\makeatletter
\newcommand{\providelength}[1]{%
  \@ifundefined{\expandafter\@gobble\string#1}
   {%
    \typeout{\string\providelength: making new length \string#1}%
    \newlength{#1}%
   }
   {%
   }%
}
\makeatother

\usepackage[pagebackref,breaklinks,colorlinks,citecolor=eccvblue]{hyperref}

\usepackage{orcidlink}

\begin{document}

\title{GaNI: Global and Near Field Illumination Aware Neural Inverse Rendering} 

\titlerunning{Abbreviated paper title}

\author{Jiaye Wu\inst{1} \and
Saeed Hadadan\inst{1} \and
Geng Lin\inst{1} \and
Matthias Zwicker\inst{1} \and
David Jacobs\inst{1} \and
Roni Sengupta\inst{2}
}

\authorrunning{J. Wu et al.}

\institute{University of Maryland, College Park, MD 20740, USA \\
\email{\{jiayewu, saeedhd, geng, zwicker, dwj\}@umd.edu} \\
 \and
University of North Carolina at Chapel Hill, NC, 27599\\
\email{ronisen@cs.unc.edu}}

\maketitle

\begin{abstract}
In this paper, we present GaNI, a Global and Near-field Illumination-aware neural inverse rendering technique that can reconstruct geometry, albedo, and roughness parameters from images of a scene captured with co-located light and camera. Existing inverse rendering techniques with co-located light-camera focus on single objects only, without modeling global illumination and near-field lighting more prominent in scenes with multiple objects. We introduce a system that solves this problem in two stages; we first reconstruct the geometry powered by neural volumetric rendering NeuS, followed by inverse neural radiosity NeRad that uses the previously predicted geometry to estimate albedo and roughness. However, such a naive combination fails and we propose multiple technical contributions that enable this two-stage approach. We observe that NeuS fails to handle near-field illumination and strong specular reflections from the flashlight in a scene. We propose to implicitly model the effects of near-field illumination and introduce a surface angle loss function to handle specular reflections. Similarly, we observe that NeRad assumes constant illumination throughout the capture and cannot handle moving flashlights during capture. We propose a light position-aware radiance cache network and additional smoothness priors on roughness to reconstruct reflectance. Experimental evaluation on synthetic and real data shows that our method outperforms the existing co-located light-camera-based inverse rendering techniques. Our approach produces significantly better reflectance and slightly better geometry than capture strategies that do not require a dark room.
\end{abstract}

\providelength\width
\setlength\width{1.3cm}

\begin{figure}
\tiny
\centering
\renewcommand{\tabcolsep}{1pt}
\begin{tabular}{cccccccccccccc}
&&&\multicolumn{3}{c}{IRON}&&\multicolumn{3}{c}{WildLight}&&\multicolumn{3}{c}{Ours} \\ 
\cmidrule{4-6} \cmidrule{8-10} \cmidrule{12-14} \\   %
{\makebox[5pt]{\rotatebox{90}{\hspace{10pt}  \tiny Input Image}}}
&\includegraphics[ width=\width]{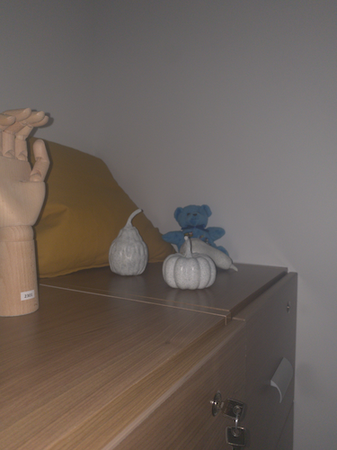}
&{\makebox[5pt]{\rotatebox{90}{\hspace{24pt}  \tiny Albedo}}}
& \includegraphics[ width=\width]{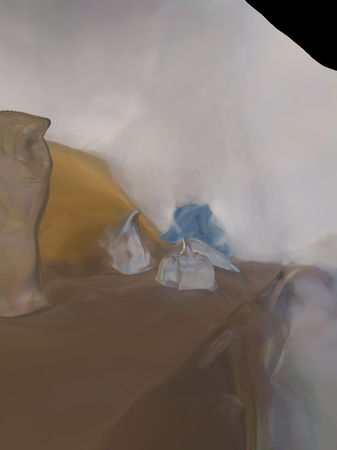}
&{\makebox[5pt]{\rotatebox{90}{\hspace{16pt}  \tiny Roughness}}}
& \includegraphics[ width=\width]{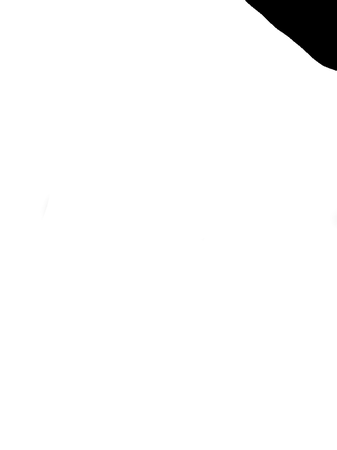}
&{\makebox[5pt]{\rotatebox{90}{\hspace{24pt}  \tiny Albedo}}}
& \includegraphics[ width=\width]{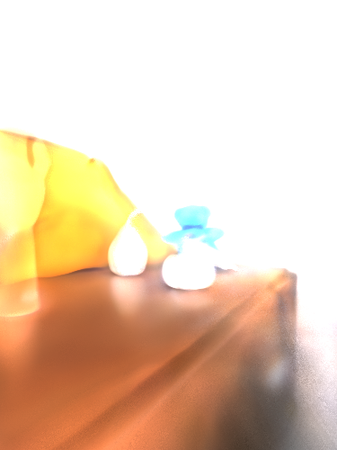}
&{\makebox[5pt]{\rotatebox{90}{\hspace{16pt}  \tiny Roughness}}}
& \includegraphics[ width=\width]{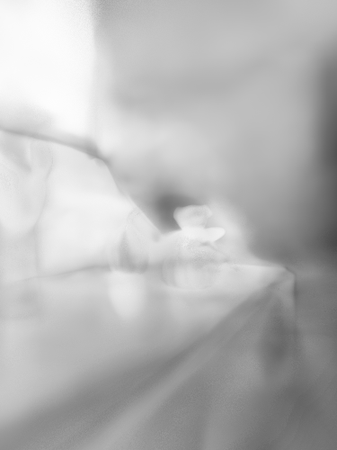}
&{\makebox[5pt]{\rotatebox{90}{\hspace{24pt}  \tiny Albedo}}}
& \includegraphics[ width=\width]{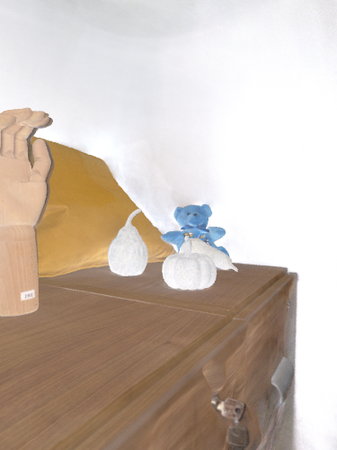}
&{\makebox[5pt]{\rotatebox{90}{\hspace{16pt}  \tiny Roughness}}}
& \includegraphics[ width=\width]{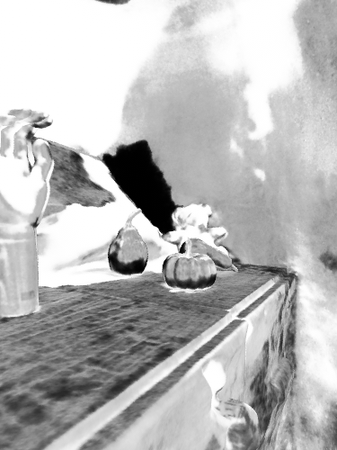} \\

&&
{\makebox[5pt]{\rotatebox{90}{\hspace{18pt}  \tiny Geometry}}}
& \includegraphics[ width=\width]{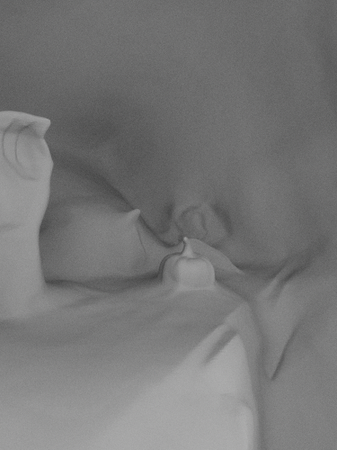}
&{\makebox[5pt]{\rotatebox{90}{\hspace{16pt}  \tiny Re-Render}}}
& \includegraphics[ width=\width]{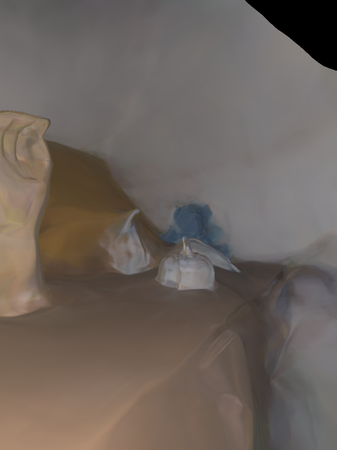}
&{\makebox[5pt]{\rotatebox{90}{\hspace{18pt}  \tiny Geometry}}}
& \includegraphics[ width=\width]{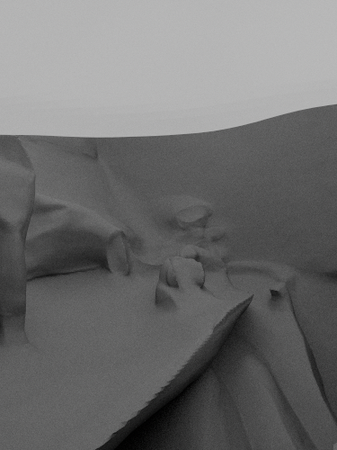}
&{\makebox[5pt]{\rotatebox{90}{\hspace{16pt}  \tiny Re-Render}}}
& \includegraphics[ width=\width]{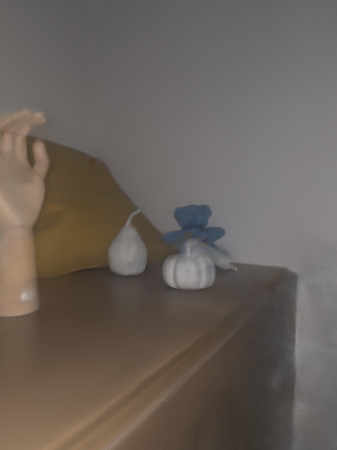}
&{\makebox[5pt]{\rotatebox{90}{\hspace{18pt}  \tiny Geometry}}}
& \includegraphics[ width=\width]{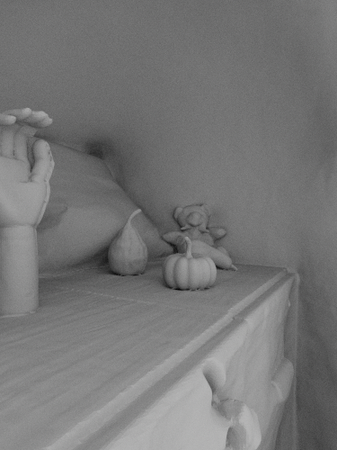}
&{\makebox[5pt]{\rotatebox{90}{\hspace{16pt}  \tiny Re-Render}}}
& \includegraphics[ width=\width]{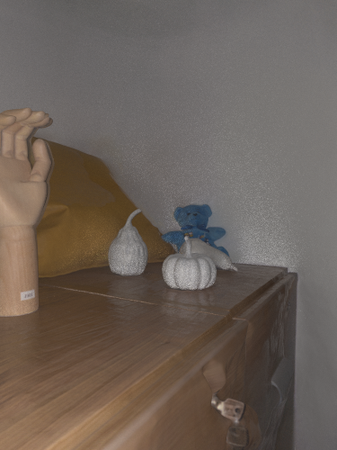}\\

\\

\end{tabular}
\caption{ We perform inverse rendering of a scene from multiple images captured with co-located light and camera. Our method, GaNI, produces better geometry, albedo, roughness and re-rendering in unseen views than state-of-the-art approaches, IRON \cite{iron} and WildLight \cite{wildlight}, that also uses co-located light-camera.
}
\vspace{-2em}
\label{fig:teaser}
\end{figure}

\vspace{-1.5em}
\section{Introduction}
\vspace{-0.5em}
Decomposing an indoor scene into geometry, material properties, and lighting, called inverse rendering \cite{deep_3d_capture, deep_reflectance,shape_material_at_home,iron,wildlight}, is a long-standing problem in computer vision. Inverse rendering has many applications in VR/AR, computational photography, and robotics perception, e.g. relighting, material editing, etc. \cite{montecarlo_indoorscene, Li_2020_CVPR, iron, wildlight}. However, existing methods are often tailored for individual objects \cite{iron,wildlight,physg,fast_light_weight_ps} or human faces \cite{sengupta2018sfsnet,kim2018inverse, total_relight}. For scenes consisting of multiple objects, recent methods have shown impressive quality in reconstructing geometry \cite{neus,geoneus}, but fail to reconstruct material reflectance i.e the Bi-directional Reflectance Distribution Function (BRDF), \cite{sengupta2019neural, Li_2020_CVPR, i2sdf, inverserendernet,irisformer}, since it is is highly under-constrained.

To make the inverse rendering problem well-constrained, researchers proposed capturing images using a co-located flashlight and camera \cite{wildlight, fast_light_weight_ps, deep_3d_capture, unified_shape_svbrdf, practical_svbrdf, iron}. Such a setup is readily accessible since one can easily capture images by turning on the flashlight of one's mobile camera; light and camera location can be easily calibrated by using Structure-from-motion which imposes additional constraints on the inverse rendering problem. Recent approaches, like IRON \cite{iron} and WildLight \cite{wildlight}, have shown impressive performance in material prediction of a single object captured using a co-located light and camera in a dark room and under ambient lighting respectively. However, these methods fail to generalize to scenes consisting of multiple objects, as noted in our experimental evaluations~\ref{fig:teaser}. We believe this is because the complexity of scenes with multiple objects surpasses that of individual objects, particularly with a co-located light-camera setup, which induces a pronounced near-field effect, where different regions in the scene receive different intensities of light. Moreover, interactions between objects and surfaces result in strong inter-reflections or global illumination.

In this paper, we aim to perform inverse rendering of a scene consisting of multiple objects captured using a co-located light and camera. We present GaNI, a system that can reconstruct both high-quality geometry and reflectance, by implicitly modeling both global and near-field illumination, hence overcoming the shortcomings of existing approaches \cite{iron,wildlight}. We focus on examples that consist of multiple objects, e.g. coffee table, shoe rack, table, etc. We believe this is a first step towards developing a system that can perform accurate inverse rendering of a 360-degree room-scale scene, which is significantly more challenging.

Our proposed method develops a neural scene representation by optimizing the parameters for each scene independently. We propose a two-stage approach where we first solve for geometry followed by reflectance, instead of jointly optimizing both which is challenging in the presence of complex illumination effects. We observe that a naive combination of first reconstructing geometry with neural volume rendering, NeuS~\cite{neus}, followed by reflectance estimation with InvNeRad~\cite{nerad}, an efficient technique to model multi-bounce global illumination with known geometry, fails due to specularity, global illumination, and changing light source. We propose several technical contributions that enable geometry and reflectance estimation using a two-stage approach involving NeuS~\cite{neus} and InvNeRad~\cite{nerad} backbones.

In the first stage, we modify NeuS\cite{neus} to handle near-field lighting and strong specular reflections which are more pronounced due to co-located light and camera. We implicitly model the near-field effect by conditioning NeuS with a parametrization that encodes the distance between the scene point from the camera. This approach enables the neural radiance representation to effectively learn and model spatially varying incident illumination fields, even in the presence of global illumination. Consequently, we achieve a robust reconstruction of scene geometry, presenting a significant improvement over NeuS\cite{neus}. To further improve reconstruction at the regions of strong specularity from the flashlight, we adaptively set weights for each point on the scene, reducing contributions from those where the light is nearly perpendicular or parallel to the surface.

In the second stage, we use surface rendering to solve the rendering equation and extract principled BRDF \cite{principled}. Here we observe that Inverse Neural Radiosity (InvNeRad) \cite{nerad}, which models 2nd-order light bounces using a learnable radiance cache, expects the scene illumination to remain static across the capture, which is not true for images captured with a co-located light and camera. We solve this by training a light position-aware radiance cache network that can handle moving light sources. While this leads to accurate albedo prediction, we note that the predicted roughness is often noisy for regions with faulty geometry, which we address using a total-variation smoothness prior over roughness.

We perform a detailed quantitative and qualitative analysis of our approach on 3 real and 4 synthetic scenes. First, we show that while co-located light and camera in a dark room imposes additional constraints on the capture, it produces significantly better reflectance than alternate capture possibilities, i.e. natural illumination capture \cite{neilfpp} and co-located flashlight under ambient lighting capture (WildLight) \cite{wildlight}. Next, we show that in comparison to the prior co-located light and camera-based inverse rendering algorithm, IRON \cite{iron}, developed for a single object, our method produces significantly better geometry and reflectance. %

In summary, our contributions are $\bullet$ We propose GaNI, a neural inverse rendering algorithm that reconstructs the geometry and reflectance of a scene consisting of multiple objects captured with a co-located light and camera. $\bullet$ We build upon the neural volume rendering technique NeuS~\cite{neus} and the neural global illumination modeling technique InvNeRad~\cite{nerad} by proposing key technical contributions such as: (i) implicitly model near-field lighting (ii) surface angle loss to downweight regions with strong specular reflections (iii) learning radiance cache with varying light source to model global illumination (iv) total-variation smoothness prior on roughness. $\bullet$ Our evaluations show that the proposed approach reconstructs significantly better geometry and reflectance compared to the state-of-the-art object-centric inverse rendering technique IRON~\cite{iron}; and predicts better reflectance and similar geometry compared to alternate capture setup, such as natural illumination~\cite{neilfpp} and WildLight \cite{wildlight}.

Our code and data will be available upon acceptance. 
\begin{figure*}[t]
    \centering
    \includegraphics[width=0.98\textwidth]{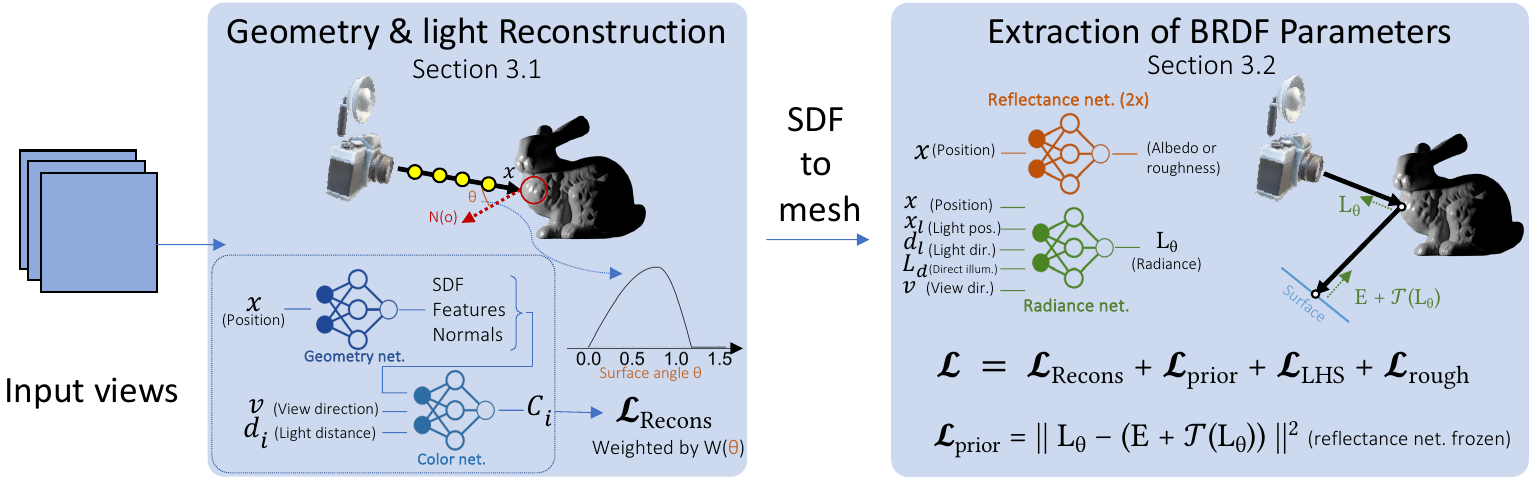}
    \caption{\textbf{Overview of our pipeline}. Our system consists of two stages. In the first stage, we reconstruct geometry with volume rendering under near-field and global illumination. In the second stage, we extract accurate material properties with surface rendering while accounting for multi-bounce global illumination while solving the rendering equation by minimizing the radiometric prior. Our output is principled BRDF\cite{principled} (albedo and roughness) represented by two separate neural networks and a neural signed distance field. }
    \label{fig:overview}
    \vspace{-2em}
\end{figure*}

\vspace{-0.5em}
\section{Related Works}
\vspace{-0.5em}
\label{related_work}

\subsubsection{Role of capture setup.} Researchers have explored different capture configurations with specific advantages that can prove beneficial depending on the intended applications. For example, some works \cite{sengupta2019neural,Li_2020_CVPR} attempt to solve inverse rendering from unconstrained images captured in-the-wild but produce low-quality results. On the other hand, researchers have used Light Stage \cite{total_relight, relightable, nlt}, a spherical gantry consisting of well-calibrated lights and cameras, to reconstruct very high-quality geometry and reflectance but can rarely be used in-the-wild. 

Recently researchers have focused on simpler, `at-home' configurations, that can be easily replicated in the wild but also provide sufficient constraints to generate better reconstructions than unconstrained approaches \cite{sengupta2019neural,Li_2020_CVPR}. These `at-home' capture setups can be under unconstraint natural illumination~\cite{nerfactor,inverse_rendering_gi, neilf, nero, physg, nvdiffrec, nvdiffrecmc}, a co-located light and camera \cite{wildlight, deep_3d_capture, unified_shape_svbrdf, practical_svbrdf, iron, colocated_handheld} or a moving camera and a flashlight \cite{shape_material_at_home,zhao2023mvpsnet}. Additionally, WildLight~\cite{wildlight} has explored inverse rendering with co-located light and camera in the presence of ambient illumination. Recent researches have shown that the choice of illumination condition can heavily impact the quality of BRDF recovery. A natural illumination capture setup is most popular due to its simplicity, as it only requires multi-view images. However, such methods often cannot extract accurate reflectance, especially in multi-object scenes with complicated geometry due to fundamental ambiguity between geometry, lighting and reflectance. On the other hand, a co-located setup often gives high quality inverse rendering results, but requires a darkroom. WildLight combines the best of the two worlds, but requires roughly equal ambient illumination energy and flashlight energy, which is difficult to satisfy in multi-object scenes. Our work is focused on co-located light and camera setup, which allows accurate inverse rendering results. We show that the state-of-the-art inverse rendering technique with co-located light and camera fails to model near-field and global illumination effects that are more visible in multi-object scenes than objects. Our approach is the first to enable co-located light-camera-based inverse rendering in scenes by implicitly modeling near-field and global illumination effects.

\vspace{-0.5em}
\subsubsection{Geometry Reconstruction.}
Recently, NeuS \cite{neus} and VolSDF \cite{volsdf} have gained popularity since they achieve high-accuracy geometry by adopting a volumetric approach to optimize a neural implicit surface representation as a Signed Distance Field (SDF). We adopt NeuS~\cite{neus} for reconstructing the geometry. However, NeuS cannot handle near-field illumination and strong specularity. Therefore, we propose to model near-field illumination and use surface angle loss to handle this scenario.

\vspace{-0.5em}
\subsubsection{Material Estimation Under Co-located Light and Camera.}
Most previous inverse rendering algorithms, such as IRON~\cite{iron}, WildLight~\cite{wildlight}, for Co-located Light and Camera do not attempt to handle global illumination. While global illumination is often negligible for objects, such inter-reflection is much more prominent for multi-object scenes. Some previous methods~\cite{inverse_rendering_gi, neilfpp} targeted at natural illumination estimate global illumination by leveraging a radiance field fitted to image observations. However such approaches are problematic for co-located light and camera setup as the radiance of only a subset of the scene can be observed. 

To properly define radiance for the entire scene requires solving the rendering equation. Such a solution typically requires solving path integrals, i.e. integration over the path space and building paths that connect the camera to the light source. The most naive approach that serves this purpose is \emph{differentiable path tracing}, which is known to have an intractable memory footprint in complex scenes and linear time complexity in the path length. To alleviate the time and memory requirements, a recent work,  InvNeRad~\cite{nerad}, adopted radiance caching techniques and only computes the primary ray intersection, and queries the cache for the contribution of the rest of the path, thus accounting for global illumination. In particular, InvNeRad \cite{nerad} uses a neural network to represent the radiance cache and solve the rendering equation in self-supervised fashion. It shows significant improvements in terms of memory and time compared to previous work for inverse rendering. 

However, InvNeRad assumes static illumination, and cannot be directly applied to co-located light and camera setup. Therefore, we propose a light source conditioned neural radiance cache to avoid creating hundreds of radiance caches and allow sharing computed radiance across different light positions. On real data, we also found small errors in geometry often lead to errors in roughness. Therefore, we propose a roughness regularization to reduce the error.

\vspace{-0.5em}
\section{Method}
\vspace{-0.5em}
\noindent \textbf{Capture Setup.} Similar to \cite{iron}, we capture images with a smartphone with a flashlight turned on in a dark room. Such a setup provides multi-illumination images, which helps resolve ambiguity between illumination and material properties that occur in a natural illumination setup, allowing our method to obtain significantly better material estimation than capture under natural illumination. While this means our method can only be used at night time, it produces significantly better BRDF without requiring any sophisticated hardware equipment or calibration.

\noindent \textbf{Overview.} We observe that existing one-stage colocated inverse rendering systems, such as WildLight~\cite{wildlight}, which represents radiance as principled BRDF (Bidirectional Reflectance Distribution Function) parameters under a co-located flashlight, are often not robust against global illumination effects, such as interreflection between objects, and fails for real geometry reconstruction.  Therefore, our reconstruction process consists of two optimization stages, as depicted in Figure \ref{fig:overview}. The first stage (Section \ref{sec:stage1}) performs volume rendering by improving NeuS~\cite{neus} to reconstruct the geometry with implicit near-field illumination and handle strong specular reflection with a surface angle loss.  The second stage (Section \ref{sec:stage2}) builds on InvNeRAD \cite{nerad} and uses the predicted geometry to solve the rendering equation at all moving light positions and extracts principled BRDF parameters\cite{principled} with a smoothness prior on roughness.

\vspace{-0.5em}
\subsection{Stage 1: Geometry Recovery with Volume Rendering}
\vspace{-0.5em}
\label{sec:stage1}
We build our volume rendering procedure on top of NeuS\cite{neus}, and we similarly represent the scene by two neural networks.

\textbf{Geometry Network} \( \operatorname{S_{\Theta_S}}(\mathbf{x}) \rightarrow \{s, \mathbf{f}\} \), which maps a 3D position \(\mathbf{x}\) to its signed distance to the closest surface, and a feature vector \(\mathbf{f}\). The network represents the geometry of the scene.

\textbf{Color Network} \( \operatorname{C_{\Theta_C}}(\mathbf{x}, \mathbf{n}, \mathbf{v}, \mathbf{f})\rightarrow \mathbf{c} \), where \(\mathbf{x}\) is the 3D position of the point, \(\mathbf{n}\) is the normal at position \(\mathbf{x}\), \(\mathbf{v}\) is the view direction of the camera and \(\mathbf{f}\) is the feature vector output by the geometry network. The network represents the radiance of the scene.

The images are rendered with volume rendering as defined by NeuS\cite{neus}. Using their unbiased and occlusion-aware weight \(\operatorname{w}(\cdot)\), with \(\mathbf{o}\) as camera position, \(\mathbf{v}\) as view direction, t as distance along camera ray, and \(\operatorname{p}(t)\) as 3D positions along camera ray, we have the following equation.

{\small
\begin{equation}
    \operatorname{C}(\mathbf{o}, \mathbf{v})=\int_0^{+\infty} w(t) \operatorname{C_{\Theta_C}}(\operatorname{p}(t), \mathbf{v}) dt
\end{equation}
}

For a larger scene, the distance from co-located light to different parts of the scene will be different. However, as shown previously, the color network of NeuS does not model such differences. As shown in table \ref{tab:synthetic_geometry_metrics} and Figure \ref{fig:qualitative_figure}, previous inverse rendering methods such as IRON~\cite{iron}, which directly use NeuS\cite{neus} often fails in our multi-object dataset.

 We model the co-located flashlight as a point light source. To handle the change of radiance, we additionally parameterize the color network with the flashlight position \(\mathbf{x_i}\):

\begin{equation}
    \operatorname{C_{\Theta_C}}(\mathbf{x}, \mathbf{n}, \mathbf{v}, \mathbf{f}, \mathbf{x_i})\rightarrow \mathbf{c}
\end{equation}

Since the camera is co-located with the flashlight, \(\mathbf{x_i} = \mathbf{x} + t \mathbf{v}\), where \(t\) is the distance between the flashlight and queried point. We can uniquely identify a \((\mathbf{x_i}, \mathbf{x}, \mathbf{v})\) combination by just \((t, \mathbf{x}, \mathbf{v})\). Additionally, since the flashlight is a point light source, the irradiance follows the inverse quadratic rule \(\frac{1}{t^2}\) as distance increases from the flashlight. As such, we propose the following parametrization:

\begin{equation}
    \operatorname{C_{\Theta_C}}(\mathbf{x}, \mathbf{n}, \mathbf{v}, \mathbf{f}, \frac{1}{t^2})\rightarrow \mathbf{c}
\end{equation}

By encouraging but not forcing the light conditioned radiance network to model near field light effects from a point light source,
we can robustly and accurately reconstruct geometry even in the presence of global illumination.

Similar to previous methods, we encode view direction with spherical harmonics\cite{instantngp} and every other input with positional encoding\cite{nerf}.

We supervise the pixels with ground-truth image pixel values via reconstruction loss. However, as our color network models the illumination change, we supervise the network in linear RGB space to bypass the non-linearity introduced by gamma function. Together, we use linearized log loss introduced by RawNeRF~\cite{rawnerf} to better learn radiance in dark regions. Denote \(y_i\) as the i-th pixel of groundtruth.

{
\small
\begin{equation}
    \label{eq:linearized_tone}
    \tilde{L}_\text{recons}(\hat{y}, y) = \sum_i \left(\frac{\operatorname{C}(\mathbf{o}, \mathbf{v})-y_i}{sg(\operatorname{C}(\mathbf{o}, \mathbf{v})) + \epsilon}\right)^2
\end{equation}
}

\textbf{Surface Angle Weighting} We observe that strong reflections from the flashlight heavily deteriorate geometry prediction. While several works, such as neural-pbir\cite{pbir} or Ref-NeuS, have explored loss weighting schemes to improve geometry reconstruction for specular objects, we found these methods often tend to degrade the geometry for concave scenes. 

Our weighting scheme is based on the fact that the light position is known, and co-located with the camera, so we can calculate the angle the light ray forms with the surface normal. Our intuition is to downweight very small angles (perpendicular direction) that produce strong specular reflections and very large angles (grazing direction) that cause the diffuse component to become minimal, while specular inter-reflection become even stronger due to fresnel effects.

Since we do not have ground truth geometry, we use the neural sdf during training as proxy geometry. We render the surface normal similar to color:

{\small
\begin{equation}
    \operatorname{N}(\mathbf{o})=\int_0^{+\infty} w(t) \nabla\operatorname{S_{\Theta_S}}(\mathbf{x}) dt
\end{equation}
}

Given angle \(\theta=\operatorname{arccos}(N\cdot L)\) between normal \(N\) and light ray orientation \(L\), we propose to weight the reconstruction loss \(L_\text{recons}\) of each pixels by:

\begin{equation}
    W_{a,b}(\theta) = \left\{
\begin{array}{ll}
      \operatorname{max}(\operatorname{cos}(a(\theta-\frac{\pi}{4})), 0) & x \leq \frac{\pi}{4} \\
      \operatorname{max}(\operatorname{cos}(b(\theta-\frac{\pi}{4})),0) & x > \frac{\pi}{4} \\
\end{array} 
\right.
\end{equation}

Our choice of function is motivated by the following requirements: decrease slowly in the middle near \(\frac{\pi}{4}\), where neither specularity nor fresnel effects are prominent, and decrease rapidly on the two extremes where those effects quickly dominate observation; the function needs to be asymmetrical, as retroreflective specular highlights are only observable from a very narrow range of angles w.r.t surface normal, but specular inter-reflection is prominent over a significantly larger range of angles. We use \(a=2\), \(b=3\) for all our experiments.

\textbf{Structure-from-Motion (SfM) pointcloud supervision on real data.} Similar to other neural reconstruction techniques, we require camera poses to be known ahead of time before reconstruction, which is commonly estimated from structure-from-motion (SfM). Similar to prior works such as Geo-NeuS\cite{geoneus}, Neilf++\cite{neilfpp}, instead of discarding the point cloud estimated in SfM, our system can optionally use the point cloud to supervise geometry by requiring the neural signed distance field to be zero at point cloud locations \(X_\text{sfm}\). We use SfM point cloud supervision on real data.

\vspace{-0.5em}
\subsection{Stage 2: BRDF Estimation with Surface Rendering}
\vspace{-0.5em}
\label{sec:stage2}

In the second stage, we assume a surface based rendering model, and by properly modeling global illumination effects, we recover material properties represented by popular principled BRDF~\cite{principled}, which allows intuitive human editing for downstream applications. Specifically, we optimize for spatially varying albedo and roughness. We represent them using separate neural networks, $\phi_a(x)$ and $\phi_r(x)$, respectively, where \(x\) is any 3D position.

With a co-located light and camera capture setup,  illumination changes in every image, and we observe a subset of the radiance field. Hence, we cannot compute global illumination directly from image observations similar to some previous inverse rendering methods\cite{inverse_rendering_gi}. To recover unobserved radiance, we leverage an efficient global illumination inverse rendering technique InvNeRad~\cite{nerad} that solves the rendering equation directly on geometry defined by the signed distance field.

\vspace{-0.5em}
\subsubsection{Background: Inverse Neural Radiosity.} InvNeRad~\cite{nerad} is an inverse rendering method that efficiently accounts for global illumination without the need to explicitly simulate multiple scattering events through path tracing. It uses a radiance cache $L_\theta$ represented as a neural network with parameter set $\theta$, and after bouncing the ray only once, it queries the cache to collect the contribution of the rest of the path. More formally, the incident radiance at pixel $k$, denoted as $I_k$, is determined by the measurement equation,
{
\small
\begin{equation}
\label{eq:RHS-measurement}
I_k = \int_{\mathcal{A}}\int_{\mathcal{H}^2} W_k(x, \omega) . (E + \mathcal{T}(L_{\theta})(x,\omega)) dx d\omega^{\perp}
\end{equation}
}
where $W_k(x,\omega)$ models the response of a sensor pixel to incident radiance over its area $\mathcal{A}$ and the hemisphere of directions $\mathcal{H}^2$, \(E\) is the emitted radiance distribution $E(x,\omega_o)$, $\mathcal{T}$ is the transport operator.

To optimize the scene parameters $\phi$ from the ground truth images, we denote the image rendered using $\phi$ as $I(\phi)$, then we can define a reconstruction loss as
{
\small
\begin{equation}
\mathcal{L}_{\mathrm{recons}}(I(\phi)) = \|I(\phi) - I^{\mathrm{GT}}\|.
\end{equation}
}

To train the radiance cache, InvNeRad~\cite{nerad} introduces a radiometric prior

{
\small
\begin{equation}
\label{eq:radiometric prior}
    \mathcal{L}_{\mathrm{prior}}(\theta) = \| L_{\theta}(x,\omega_o) - (E(x,\omega_o) + \mathcal{T}(L_{\theta})(x,\omega_o))\|.
\end{equation}
}

Following InvNeRad~\cite{nerad}, we additionally constrain the neural radiance cache with groundtruth radiance from images by directly rendering the neural radiance cache into an image \(I^{LHS}_k\). Then we define an additional loss term to match the image to groundtruth.
{
\small
\begin{align}
    \mathcal{L}_{\mathrm{LHS}}(\theta)= \left\|I^{LHS}(\theta) - I^{\mathrm{GT}} \right\|^2.
    \label{eq:lhs-recons-loss}
\end{align}
}

\vspace{-0.5em}
\subsubsection{Co-located Light Inverse Neural Radiosity.}
InvNeRad assumes illumination is constant across the input views. With our co-located light capture setup, the lighting changes with each view. A naive approach is to have a radiance cache for every flashlight \(l\), and formulate the radiometric prior as:

{
\small
\begin{equation}
\label{eq:colocated_radiometric_prior}
    \mathcal{L}_{\mathrm{prior}}(\theta) = \| L_{l,\theta}(x,\omega_o) - (E_l(x,\omega_o) + \mathcal{T}(L_{l,\theta})(x,\omega_o))\|
\end{equation}
}
where for each captured image with flashlight $l$, $L_{l, \theta}$ requires a separate network. This is computationally prohibitive as we could easily have close to 1000 different light positions, thus 1000 different networks. Instead, we propose to use a single radiance cache network to represent radiance at all flashlight positions, which will take additional parameters \(x_l\) and \(d_l\), the position of the flashlight and orientation, as input. We also noticed that under a co-located light scenario, direct illumination often has strong discontinuities that change with light positions, making it hard for the network to model. However, it can be computed easily. Therefore, we additionally condition the neural radiance network on direct illumination \(L_\text{direct}\). Denote visibility of surface from flashlight as \(\mathbbm{1}_\text{vis}\),  flashlight radiant intensity as \(E_\text{flash}\), and distance between flashlight and surface as \(d\).

{
\small
\begin{equation}
    L_\text{direct}= \mathbbm{1}_\text{vis} \frac{E_\text{flash} }{d^2}
\end{equation}
}

\vspace{-0.5em}
\subsubsection{Roughness Regularization.}
Unlike the original InvNeRad setup, we do not have access to groundtruth geometry. We found small imperfections will often lead to significant errors in roughness. Therefore, we additionally incorporate a total variation loss for roughness, and we set \(\lambda = 0.02\) in all our experiments.

{
\small
\begin{equation}
    L_{rough} = \lambda |\nabla \phi_r|
\end{equation}
}

Finally, the optimization task is
{
\small
\begin{equation}
\label{eq:objective}
\phi^*, \theta^* = \argmin_{\phi,\theta} \mathcal{L}_{\mathrm{recons}}(I(\phi)) + \mathcal{L}_{\mathrm{prior}}(\theta) + \mathcal{L}_{\mathrm{rough}}(\phi) + \mathcal{L}_{\mathrm{LHS}}(\theta)
\end{equation}
}

\vspace{-0.5em}
\subsubsection{Implementation Details}
Similar to InvNeRad, we implement our second stage on top of Mitsuba 3~\cite{mitsuba3}, and use hashgrid encoding to encode the 3D position inputs to the reflectance fields $\phi_a$ and $\phi_r$. We also use the hashgrid encoding for flashlight positions, and spherical harmonics for flashlight orientation to encourage smoothness between contiguous flashlight positions but still allow high frequency changes.

To further speedup global illumination computation, we extract the neural sdf at $2048^3$ resolution and store as a mesh. We found the resolution sufficient to represent the details of the scenes we tested.

\begin{table}[!h]
\vspace{-1em}
\tiny
\centering
\caption{\textbf{Quantitative comparison of geometry on synthetic data.} Our method produces significantly better geometry than IRON \cite{iron} and slightly better geometry than WildLight \cite{wildlight} with images captured with co-located lighting and camera in a darkroom.
}
\renewcommand{\tabcolsep}{2pt}
\begin{tabular}{rcccccccc}
& & \multicolumn{3}{c}{\tiny Chamfer ($\times 100$)} & & \multicolumn{3}{c}{\tiny Depth Map L1} \\
\cmidrule{3-5} \cmidrule{7-9}
& & {\tiny Bedroom}  & {\tiny Shelf}  & {\tiny Counter} 
& & {\tiny Bedroom}  & {\tiny Shelf}  & {\tiny Counter} 
\\
\cmidrule{3-9}
IRON & &
1.012 & 0.194 & 1.672 & &
0.124 & 0.076 & 0.174  \\
WildLight & &
\textbf{0.000} & \underline{0.013} & \underline{0.076} &&
\textbf{0.004} & \underline{0.026} & \underline{0.024} \\
Ours & &
\underline{0.032} & \textbf{0.011} & \textbf{0.023}   & &
\underline{0.012} & \textbf{0.014} &  \textbf{0.015} \\
\end{tabular}

\vspace{-2em}
\label{tab:synthetic_geometry_metrics}
\end{table}

\vspace{-0.5em}
\section{Evaluation}
\vspace{-0.5em}

We perform qualitative evaluation on real data and quantitative and qualitative evaluation on synthetic data.

\vspace{-0.5em}
\subsubsection{Synthetic data.} We created three scenes, \textbf{bedroom}, \textbf{shelf}, and \textbf{kitchen counter} based on openly available Blender and Mitsuba scenes. We render the dataset with a moving co-located point light and camera as HDR images with the Mitsuba 3 renderer. Each synthetic scene contains 1000 training images and 50 validation images with randomly generated camera poses. 

\noindent  \textbf{Real data.} We captured four real scenes using co-located light and camera, \textbf{window sill}, \textbf{table}, \textbf{shoe rack}, and \textbf{coffee table} that contain 1006, 704, 564, and 650 images respectively. We reserve 5\% of the images of each scene as the validation split. %

\begin{figure}[tbh]
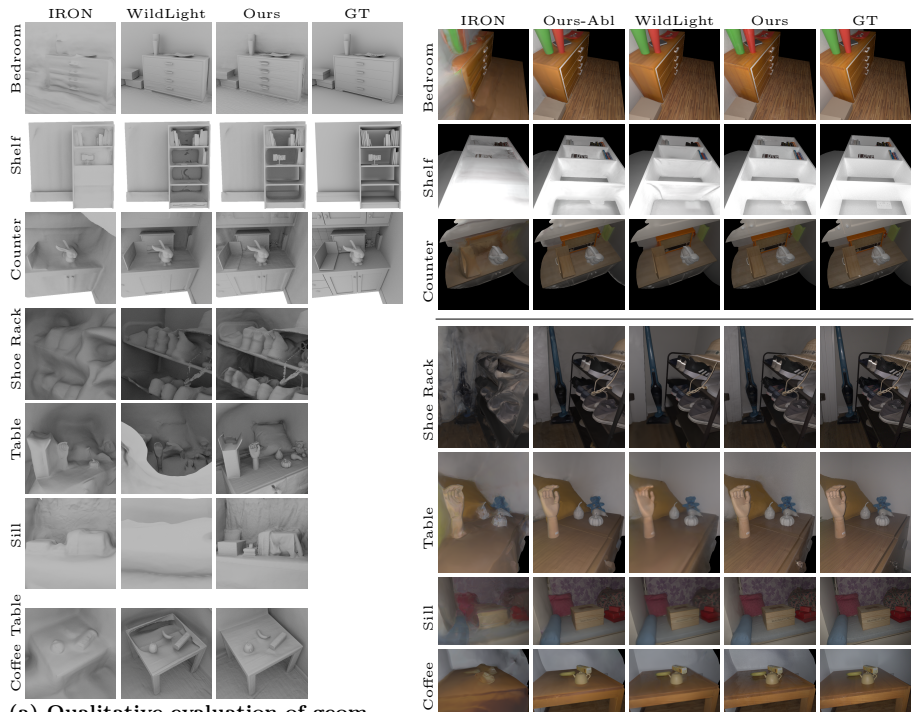

\renewcommand{\tabcolsep}{1pt}

\providelength\width
\setlength\width{1.2cm}
\begin{subfigure}{0.4\textwidth}
\tiny
\centering
\begin{tabular}{ccccc}
& IRON & WildLight & Ours & GT \\
\input{generated/geometry_figure_generated}
\end{tabular}
\vspace{-10pt}
\caption{\textbf{Qualitative evaluation of geometry.} Our method produces significantly better geometry than IRON \cite{iron} on synthetic and real data. Compared to WildLight \cite{wildlight} we produce slightly better geometry on synthetic data and significantly better on real data.}
\label{fig:geom_comparison}
\end{subfigure}
\renewcommand{\tabcolsep}{1pt}
\providelength\width
\setlength\width{1.2cm}
\begin{subfigure}{0.6\textwidth}
\tiny
\centering
\begin{tabular}{cccccc}
& IRON & Ours-Abl & WildLight & Ours & GT  \\
\input{generated/qualitative_generated_val}
\end{tabular}
\caption{\textbf{Qualitative comparison of re-rendering.} We present re-rendering in validation views for the synthetic scenes (row 1-3) and real scenes (row 4-7). Our method produces better re-rendering w.r.t. IRON~\cite{iron} due to our ability to better model near-field and global illumination.}
\label{fig:qualitative_figure_rerender}
\end{subfigure}
\vspace{-2em}
\caption{Qualitative comparison of geometry and re-rendering.}
\vspace{-2em}
\end{figure}

\begin{table}[]
\vspace{-1em}
    \centering
    \tiny
        \caption{\textbf{Qualitative evaluation of reflectance on synthetic data (Stage 2)}. We present MSE ($\times 10$) of re-rendering in unseen views, albedo, and roughness on the validation set of synthetic data. Best is shown in \textbf{bold} and second places \underline{underlined}. } 
    \renewcommand{\tabcolsep}{0.5pt}

    \begin{tabular}{rcccccccccccccccccc}
 &  & \multicolumn{4}{c}{Validation View} & & \multicolumn{4}{c}{Albedo} & & \multicolumn{4}{c}{Roughness} \\
 \cmidrule{3-6} \cmidrule{8-11} \cmidrule{13-16}
&  & Bedroom  & Shelf & Counter & \textit{Mean} &  & Bedroom  & Shelf & Counter & \textit{Mean} &  & Bedroom  & Shelf & Counter & \textit{Mean} \\
\cmidrule{3-16}
IRON &  & 0.122 & 0.212 & 0.029 & 0.121 &  & 0.565 & \underline{0.278} & \underline{0.271} & \underline{0.371} &  & 1.003 & 2.530 & 2.645 & 2.059 \\

WildLight &  & \textbf{0.006} & \underline{0.040} & \underline{0.005} & \underline{0.017} &  & 0.101 & 1.158 & 0.603 & 0.621 &  & \underline{0.636} & \underline{1.340} & \underline{1.933} & \underline{1.303} \\

\cmidrule{3-16}
Ours &  & \underline{0.010} & \textbf{0.023} & \textbf{0.005} & \textbf{0.013} &  & \textbf{0.034} & \textbf{0.051} & \textbf{0.036} & \textbf{0.041} &  & \textbf{0.380} & \textbf{1.276} & \textbf{1.118} & \textbf{0.924} \\

Ours-Abl &  & 0.016 & 0.046 & 0.014 & 0.025 &  & \underline{0.079} & 1.404 & 0.665 & 0.716 &  & 0.677 & 2.656 & 2.662 & 1.998 \\

    \end{tabular}
    \label{tab:main_metrics}
    \vspace{-4em}
\end{table}
\providelength\width
\setlength\width{1.1cm}
\begin{figure*}[!t]
\tiny
\centering

\renewcommand{\tabcolsep}{1pt}
\begin{tabular}{cccccccccccccc}
& \multicolumn{5}{c}{Albedo} && \multicolumn{5}{c}{Roughness} \\
\cmidrule{2-6} \cmidrule{8-12} 
& IRON & Ours-Abl & WildLight & Ours & GT && IRON & Ours-Abl & WildLight & Ours & GT \\

\input{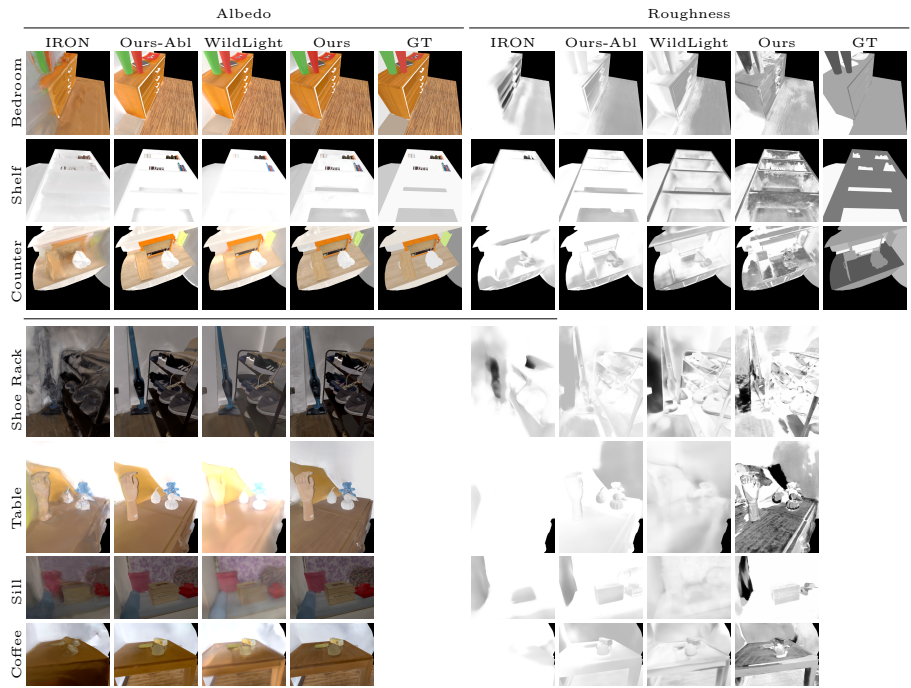}

\end{tabular}
\vspace{-10pt}
\caption{
\textbf{Qualitative comparison of reflectance estimation.} We present estimated albedo, and roughness in validation views for the synthetic scenes (row 1-3) and real scenes (row 4-7). Our method produces significantly better albedo, roughness and re-rendering w.r.t. IRON~\cite{iron} due to our ability to better model near-field and global illumination. \textbf{(Please zoom in for better visualization)} }
\label{fig:qualitative_figure}
\vspace{-2em}
\end{figure*}

\vspace{-0.5em}

\subsubsection{Comparison.} We compare our approach with that of WildLight~\cite{wildlight}, which does not handle global illumination, and IRON~\cite{iron}, which additionally does not model near-field illumination. We test all compared methods in a darkroom. While WildLight is designed for capture under ambient ilumination, we found such design has trouble converging correctly for multi-object scenes under ambient illumination. We will show the results in supplementary. The original implementation of IRON uses Mitsuba roughplastic BRDF. Since all of our synthetic scenes use principled BRDF, we modified IRON to use principled BRDF for a fair comparison. We also compare with Nelif++~\cite{neilfpp} which performs inverse rendering under natural illumination without requiring a dark room.

We reconstruct better geometry and reflectance than IRON~\cite{iron} and WildLight~\cite{wildlight} (see Fig. \ref{fig:geom_comparison}, \ref{fig:qualitative_figure} and Tab. \ref{tab:synthetic_geometry_metrics}, \ref{tab:main_metrics}). Compared to other capture setups that do not require dark room, Nelif++~\cite{neilfpp} we improve the geometry slightly but significantly improve BRDF prediction.

\vspace{-0.5em}
\subsection{Comparison with Inverse Rendering Techniques that use Co-located Light \& Camera}
\vspace{-0.5em}

\subsubsection{Evaluation of Geometry.}
We compare the geometry reconstructed by our method with state of art inverse rendering methods using a co-located capture setup. We compare to  IRON~\cite{iron} and WildLight\cite{wildlight} quantitatively on synthetic data in Tab. \ref{tab:synthetic_geometry_metrics} and qualitatively on both synthetic and real data in Fig. \ref{fig:geom_comparison}. 

As the mesh models of our synthetic scene contain hidden parts not visible in any camera views, we cannot naively compute the distance between reconstructed geometry and ground-truth meshes. Therefore, we rendered a depth map of ground-truth mesh and reconstructed mesh on validation views, and compute the L1 distance between the two depth maps as our depth map score. Additionally, we can convert the depth map of ground-truth mesh and of reconstructed mesh as two-point clouds. Then we compute the Chamfer distance from the ground-truth point cloud to the point cloud of the reconstructed mesh.

IRON~\cite{iron} does not attempt to model near-field illumination and performs poorly in multi-object scenes. WildLight~\cite{wildlight} assumes a point light source model which does not model global illumination. Therefore, while WildLight performed well on an open scene without much inter-reflection, such as `bedroom', the performance is worse on scenes with strong concavity such as `shelf', `kitchen counter' and complex geometry, e.g. the sign in `shelf' scene, the teddy bear in the back of the kitchen `counter', or the left-most shoe on the second row of the `shoe rack'. Additionally, WildLight often catastrophically fails on many
real scenes with prominent inter-reflection, such as the `table', `window sill', or `coffee table'. %

In comparison, our method is more robust in the presence of global illumination. However, since we do not strictly enforce a physically based BRDF model but implicitly regularize the radiance with a neural network, we perform slightly worse in textureless areas such as the wall.

\providelength\width
\setlength\width{1.2cm}
\begin{wrapfigure}{R}{0.5\linewidth}
\tiny
\centering
\renewcommand{\tabcolsep}{1pt}
\begin{tabular}{ccccc}
& Image & Re-Render & albedo & roughness \\
\input{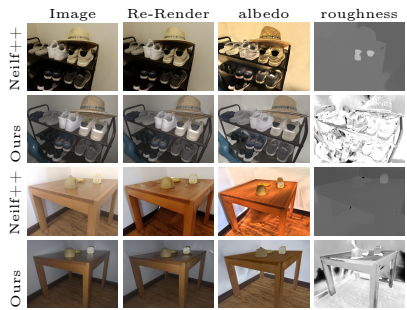} 
\end{tabular}
\caption{
Comparison of our method with Neilf++~\cite{neilfpp}, a state-of-the-art inverse rendering algorithm for natural illumination. We captured the same scene under both natural and co-located illuminations with similar number of images and camera poses. We found that our method significantly outperforms Neilf++, epsecially in albedo.
}
\vspace{-2em}
\label{fig:natural_illumination_comparison}
\end{wrapfigure}

\vspace{-0.5em}
\subsubsection{Quantitative Evaluation of Reflectance on Synthetic data}
We compare the reflectance estimation of our method with that of IRON~\cite{iron} quantitatively on synthetic data in \ref{tab:main_metrics} and qualitatively on both synthetic and real data in Figure \ref{fig:qualitative_figure} on held-out validation set. Along with albedo and roughness estimation, we also re-render the scene in validation views, unseen during training, to visualize the overall effect of inverse rendering. We compute Mean Squared Error (MSE) for re-rendering in validation views and prediction of albedo and roughness. During re-rendering in validation views, we clip the range of both ground-truth and prediction to \([0, 1]\) to prevent high-intensity pixels from dominating the error.

For quantitative evaluation, we optimize the point flashlight energy during training for all methods. Therefore, we rescale the predicted albedo of each method by the flashlight energy. For qualitative evaluation, we mask out pixels that do not have the corresponding geometry reconstructed.

Overall, our method significantly outperforms IRON~\cite{iron} on average for albedo estimation (MSE 0.066 vs 0.552), roughness estimation (MSE: 0.846 vs 1.792), and re-rendering in validation views (MSE: 0.014 vs 0.129). Similarly on real data, we significantly outperform IRON~\cite{iron}. Re-rendering images of IRON~\cite{iron} contain many artifacts due to poorly reconstructed geometry, e.g. Cabinet gets fused with in wall in Bedroom; two layers of `shelf' are reconstructed as one solid block. On the other hand, our method has very minimal errors in re-rendering of novel views, and hence predicted geometry and reflectance.  %

\vspace{-0.5em}
\subsection{Comparison with Natural Illumination and Hybrid Illumination}
\vspace{-0.5em}
Both our approach and IRON expects dark room for capture, which means it can be only performed at nighttime. Recent methods like Neilf++\cite{neilfpp} can perform inverse rendering in the presence of natural illumination. that our method outperforms Neilf++\cite{neilfpp}, which can capture scenes under natural illumination. 

We captured the same scene with similar number of images and poses under both natural and co-located illumination. We captured 750 images under both illuminations for the first example shoe rack, and 649 and 686 for co-located and natural illumination respectively for the second example coffee table. However, in Fig. \ref{fig:natural_illumination_comparison} we show prominent errors exist on the right side of the white shoe under the hat, as incorrect estimation of local illumination results in very high intensity albedo. Shading also gets baked into albedo of the hat. Similarly shading are baked into the albedo of the table for the second example. The difference is more obvious for reflectance estimation, and hence generating re-rendering views, highlighting the importance of using co-located capture for high-quality reflectance estimation.

\providelength\width
\setlength\width{2.8cm}
\begin{wrapfigure}{R}{0.5\linewidth}
\tiny
\centering
\renewcommand{\tabcolsep}{1pt}
\begin{tabular}{cc}
Ours & w/o our weighting \\
\includegraphics[width=\width]{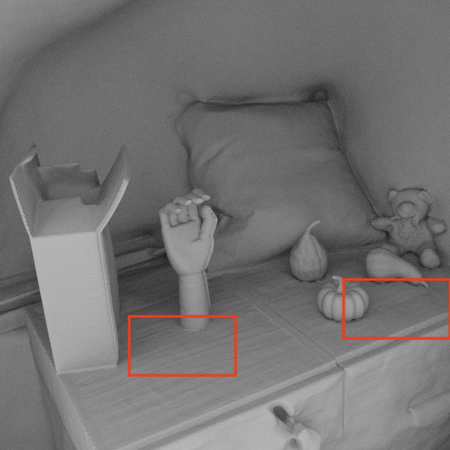}
& \includegraphics[width=\width]{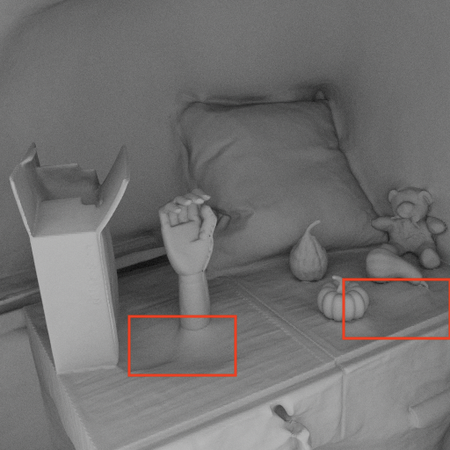}
\\

\end{tabular}
\captionof{figure}{
Ablation of our proposed surface angle weighting. Red box highlights areas where specular inter-reflection causes artifacts in geometry without surface angle weighting. Such error become significantly less pronounced in our full variant.}
\vspace{-2em}
\label{fig:loss_abl}
\end{wrapfigure}

\vspace{-0.5em}
\subsection{Ablation studies}
\vspace{-0.5em}
We perform ablation of individual component of our system, surface angle weighting, and our second stage to show their effectiveness. %

\subsubsection{Ablation study of Surface Angle Weighting}
In Figure \ref{fig:loss_abl}, we conduct ablation study of our proposed weighting scheme in the first stage. The artifacts on the table highlighted in red are present in the ablated version which is caused by interreflection. Such artifacts become significantly less pronounced in the full version as we put less weight on pixels at parallel angle.

\subsubsection{Ablation study of Ours Second Stage}
To separate the effect of geometry and reflectance, we perform an ablation study where we replace the second stage of our system with IRON's second stage, which we denote as Ours-Abl. Ours-Abl uses the same reconstructed geometry, only differing from ours in material properties extraction. As shown in Figure \ref{fig:qualitative_figure}, even with good geometry initialization, Ours-Abl often leaves strong artifacts in the albedo, such as on the side of cabinet of \textbf{kitchen counter} or the top right corner of the pillow in \textbf{table}. Without proper handling global illumination, Ours-Abl often heavily bakes global illumination effects into albedo. Moreover, Ours-Abl often has trouble recovering roughness in regions where global illumination is prominent, such as the side of cabinet, or the inside of shelf.

\vspace{-0.5em}
\section{Conclusion}
\vspace{-0.5em}
We introduce a novel system for inverse rendering for scenes using co-located light and camera. By modeling changing point light source near field illumination, we are able to obtain robust and accurate reconstruction of geometry under co-located flashlight. Then we propose light position conditioned radiance cache which extends InvNeRad to co-located light and extracts accurate material properties under global illumination. Our experiments show strong performance against state of the art methods both in geometry reconstruction and reflectance recovery. We believe our work will be an important step toward accurate inverse rendering of scenes.

\noindent\textbf{Limitations} Similar to previous co-located light and camera approaches, our approach requires capturing at nightime with lights off. Our current approach is focused on reconstruction of multi-object scenes rather than room scale scenes. One potential future direction is to extend our approach to full room scale scenes.

\section*{Acknowledgement}
Zwicker, Hadadan and Lin were supported in part by NSF grant no. 2126407. This research is based in part upon work supported by the Office of the Director of National Intelligence (ODNI), Intelligence Advanced Research Projects Activity (IARPA), via IARPA R\&D Contract No. 140D0423C0076. The views and conclusions contained herein are those of the authors and should not be interpreted as necessarily representing the official policies or endorsements, either expressed or implied, of the ODNI, IARPA, or the U.S. Government. The U.S. Government is authorized to reproduce and distribute reprints for Governmental purposes notwithstanding any copyright annotation thereon.

\clearpage  %
\bibliographystyle{splncs04}
\bibliography{main}

\setcounter{page}{1}

\setcounter{section}{0}
\renewcommand{\thesection}{A\arabic{section}}

\section{Summary}
\begin{figure*}
\tiny
\centering

\renewcommand{\tabcolsep}{1pt}
\begin{tabular}{cccccc}

& Image & Re-Render & Albedo & Roughness & Geometry  \\

\input{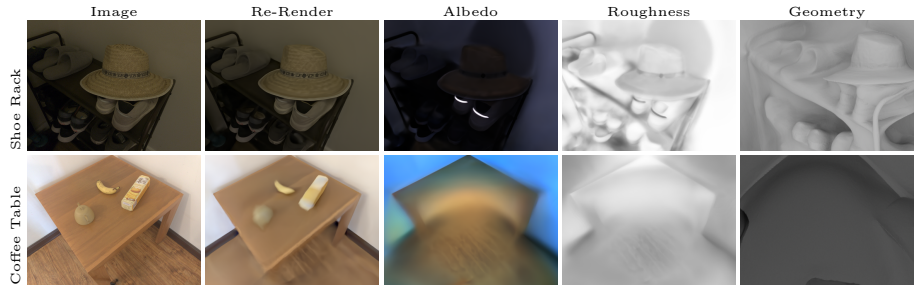}

\end{tabular}
\vspace{-10pt}
\caption{
Qualitative results of WildLight on Co-located Light and Camera under ambient natural illumination.
}
\label{fig:suppl_wildlight}
\end{figure*}

Here we summarize our supplementary. 
\begin{itemize}
    \item We render all our scenes under natural lighting conditions with our recovered geometry and material parameters using path tracing renderer mitsuba~\cite{mitsuba3}, and denoise with its built-in NVIDIA OptiX denoiser~\cite{optix}. The videos feature ambient lighting with moving point light sources and cameras to demonstrate the practical applications of our method. They are included as separate mp4 files.
    \item In Figure~\ref{fig:suppl_wildlight}, we show qualitative results of WildLight on WildLight style capture setup (Co-Located Light and Camera under ambient natural illumination). We found WildLight often fail to converge on our multi-object scenes.
    \item In section~\ref{sec:data_processing}, we provide additional details regarding how we capture and process our real data.
    \item In section~\ref{sec:stage1}, we provide additional details for our stage 1 architecture and training.
    \item In section~\ref{sec:stage2}, we provide additional details for our stage 2 training.
    \item In Figure~\ref{fig:suppl_qualitative_figure_synthetic_1}, Figure~\ref{fig:suppl_qualitative_figure_synthetic_2}, Figure~\ref{fig:suppl_qualitative_figure_synthetic_3}, we 
provide additional visualization of qualitative comparisons on synthetic data. In Figure~\ref{fig:suppl_qualitative_figure_real_1}, Figure~\ref{fig:suppl_qualitative_figure_real_2}, Figure~\ref{fig:suppl_qualitative_figure_real_3}, Figure~\ref{fig:suppl_qualitative_figure_real_4}, we provide additional visualization of qualitative comparisons on real data. Additional details of the comparison are discussed in section~\ref{sec:qual_vis}.
\end{itemize}
\providelength\width
\setlength\width{2.3cm}
\section{Real Data Capture Setup and Post-processing}
\label{sec:data_processing}
We capture all of our real data using an iPhone XS Max and an iPhone 11 Pro. We capture all the image with ProCamera app on iOS as raw dng file. During capture, we keep manual and fixed white balance, focus and exposure. For co-located capture, we also keep flashlight constantly on through the capture session. We process the raw files with RawPy~\cite{rawpy}, which is a python interface around libraw~\cite{libraw}. We perform structure-from-motion reconstruction using pixel perfect sfm~\cite{pixsfm}. We apply camera undistortion parameters estimated by pixel perfect sfm to our captured images. We found that both iPhone XS Max and iPhone 11 Pro experience significant vignetting. To calibrate for vignetting, we use a piece of white paper on a sunny day under direct sunlight as the calibration target. We model the vignetting as 6-th degree even order polynomial~\cite{vignetting}, and apply vignetting correction accordingly. We store all final processed images as 16-bit unsigned png images with linear response curve without any gamma curve applied, which are used for all following experiments.

\section{Additional Details of Stage 1 Architecture and Training}
\label{sec:stage1}
In section~3.1 of the main paper, we described some changes to original NeuS architecture, including introducing light position conditioned radiance field. Here we described additional details of implementation. 

With our synthetic dataset, the images can contain ``background pixels'' where the primary ray from the camera does not intersect with any scene geometry during rendering. Since we are not using environment map for our scenes, the values of these pixels are undefined, and we use a per image binary mask to ignore these pixels during training. Consequently, we do not put any supervision in the background region. To prevent the network from producing arbitrary values for the background, we adopt mask loss commonly used by prior works~\cite{neus,idr}. NeuS~\cite{neus} defines unbiased weights \(w_{k,i}\) along the k-th camera ray based on the underlying signed distance field. Denote \(\hat{O}_k=\sum_{i=1}^nw_i\) as the the sum of weights along the k-th camera ray, \(M_k = \{0, 1\}\) as the value of the binary mask on the k-th pixel, and \(\operatorname{BCE}\) as the binary cross entropy loss, we have the following equation.

\begin{equation}
    L_\text{mask}=\operatorname{BCE}(M_k, \hat{O}_k)
\end{equation}

Such mask loss is only used for synthetic data, and not used for real data.

Stage 1 is trained with batch size of 512 rays, learning rate of \(5\times10^{-4}\) for 500K steps on synthetic data, and 1M steps on real data.

\section{Additional Details of Stage 2 Architecture and Training}
\label{sec:stage2}
Here we provide additional details regarding the second stage of our system. We use Principled BRDF~\cite{principled} in stage 2. Similar to InvNeRad\cite{nerad}, we only optimize for albedo and roughness, while keeping other parameters fixed. We set all fixed parameters to zeros, except for ``specular'' (which sometimes called ``specular albedo''), which we set to 0.6 for synthetic scenes and 0.5 for real scenes. For fairness of comparison, we also fix ``specular'' to the same values for IRON\cite{iron}.

We train our stage 2 for \(48000\) iterations at learning rate of \(5\times10^{-4}\)  with batch size of 16384 for synthetic data and 15625 for real data.
\section{Additional Visualization of Qualitative Results} 
\label{sec:qual_vis}
\begin{figure*}
\tiny
\centering

\renewcommand{\tabcolsep}{1pt}
\begin{tabular}{cccccccc}

& IRON & Ours-Abl & WildLight & Ours & GT \\

\input{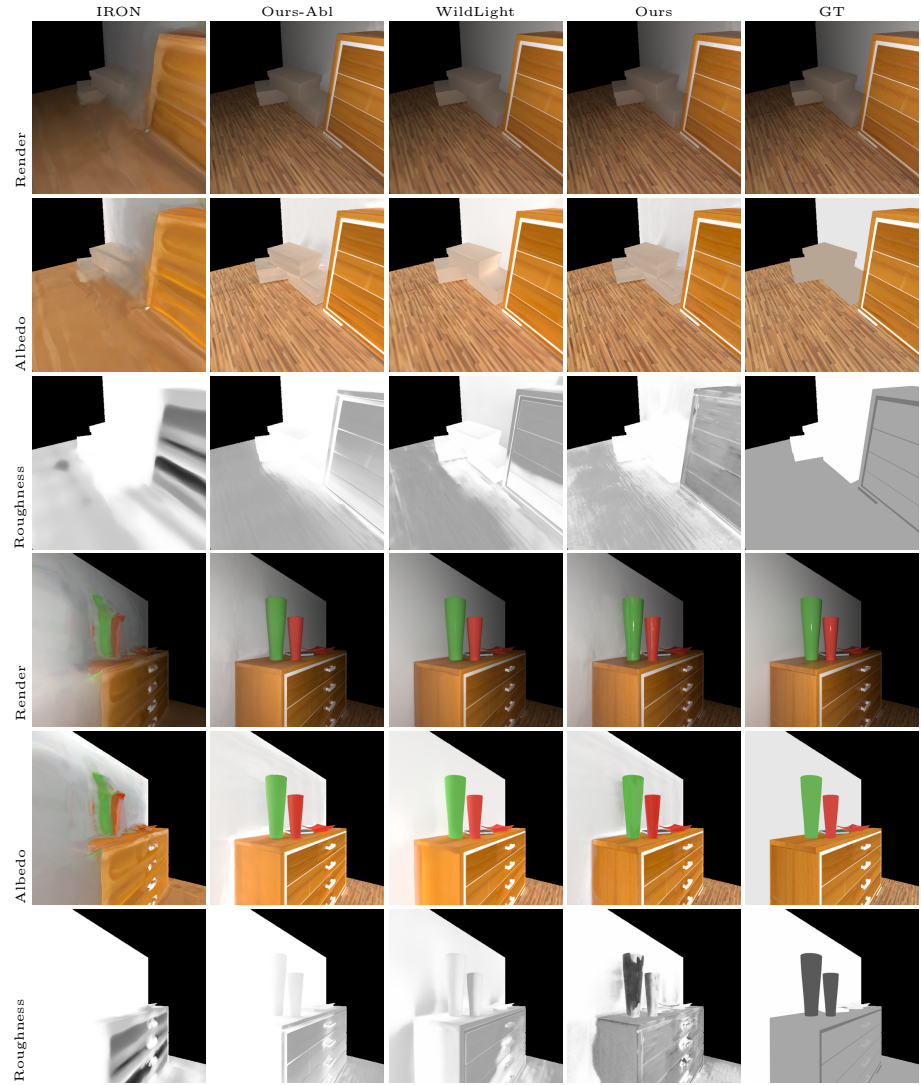}

\end{tabular}
\vspace{-10pt}
\caption{
Qualitative comparison on synthetic scene bedroom. 
}
\label{fig:suppl_qualitative_figure_synthetic_1}
\end{figure*}

\begin{figure*}
\tiny
\centering

\renewcommand{\tabcolsep}{1pt}
\begin{tabular}{cccccccc}

& IRON & Ours-Abl & WildLight & Ours & GT \\

\input{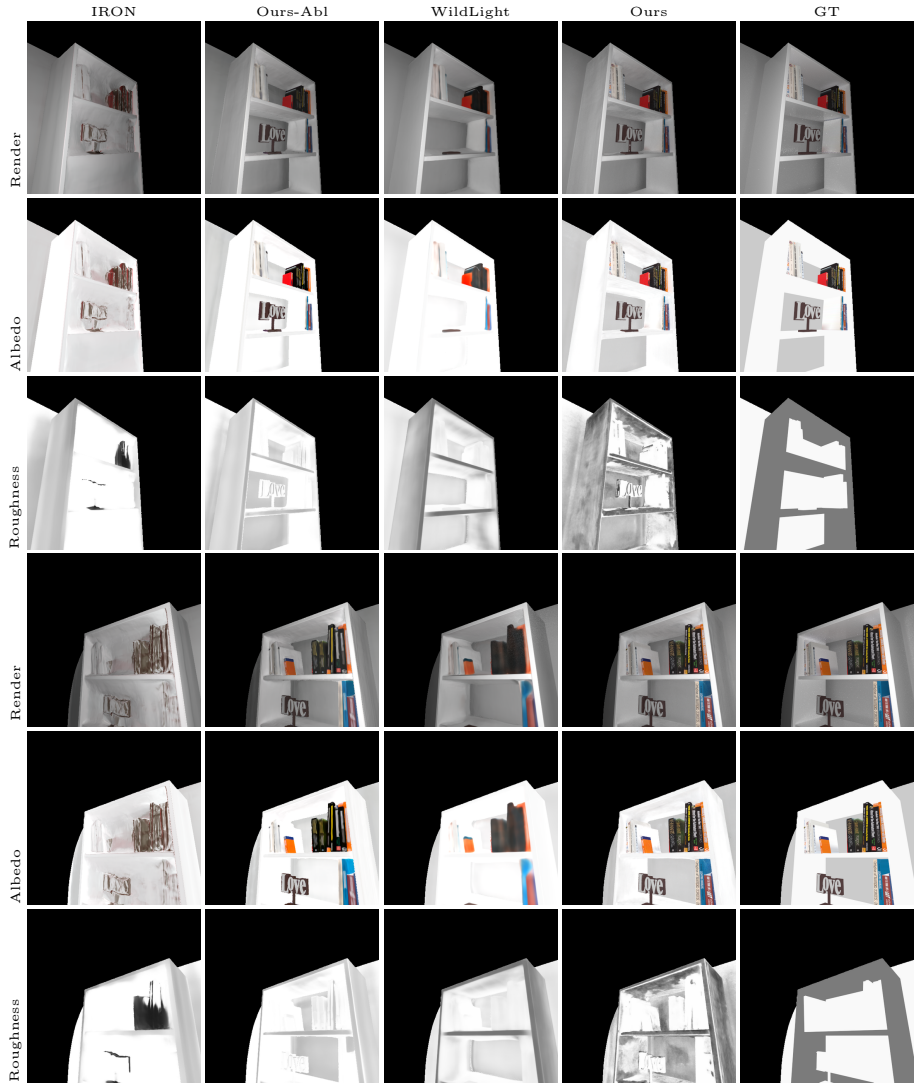}

\end{tabular}
\vspace{-10pt}
\caption{
Qualitative comparison on synthetic scene coffee table. 
}
\label{fig:suppl_qualitative_figure_synthetic_2}
\end{figure*}

\begin{figure*}
\tiny
\centering

\renewcommand{\tabcolsep}{1pt}
\begin{tabular}{cccccccc}

& IRON & Ours-Abl & Ours & GT \\

\input{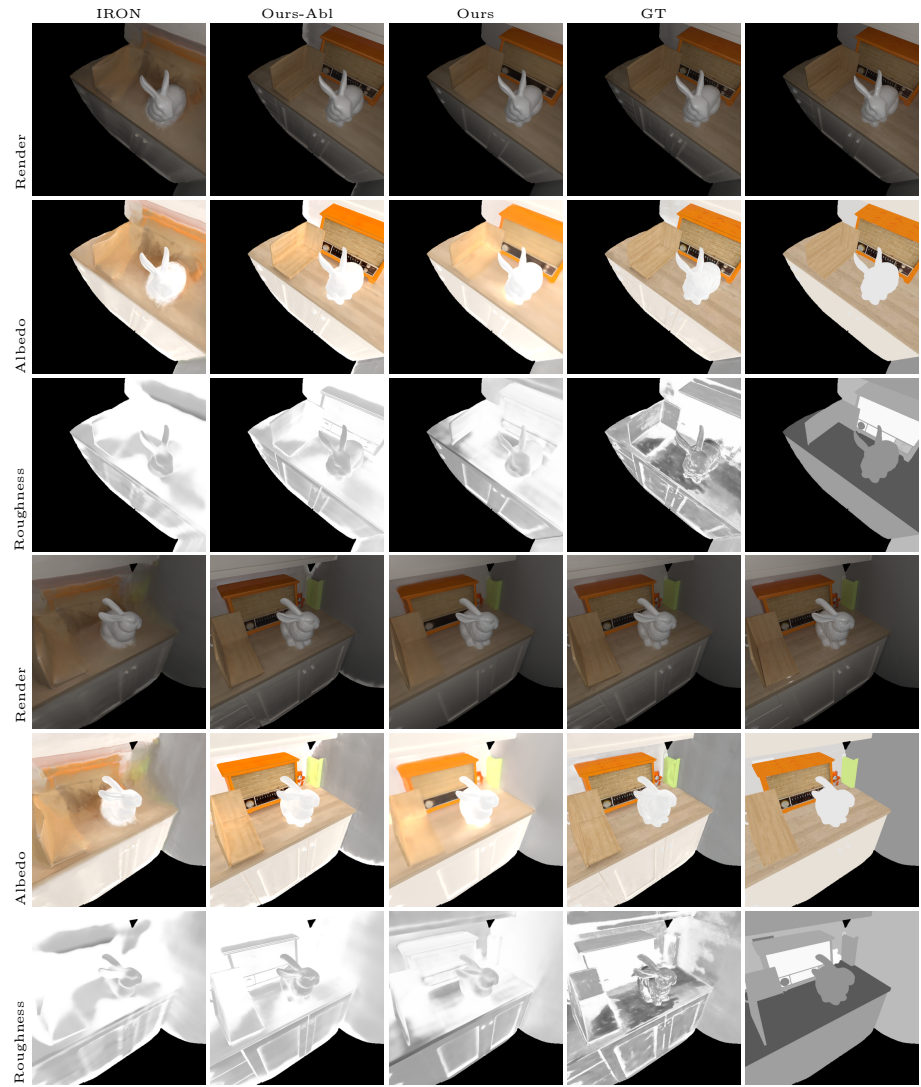}

\end{tabular}
\vspace{-10pt}
\caption{
Qualitative comparison on synthetic scene shelf.
}
\label{fig:suppl_qualitative_figure_synthetic_3}
\end{figure*}

\begin{figure*}
\tiny
\centering

\renewcommand{\tabcolsep}{1pt}
\begin{tabular}{cccccccc}

& IRON & Ours-Abl & WildLight & Ours & GT \\

\input{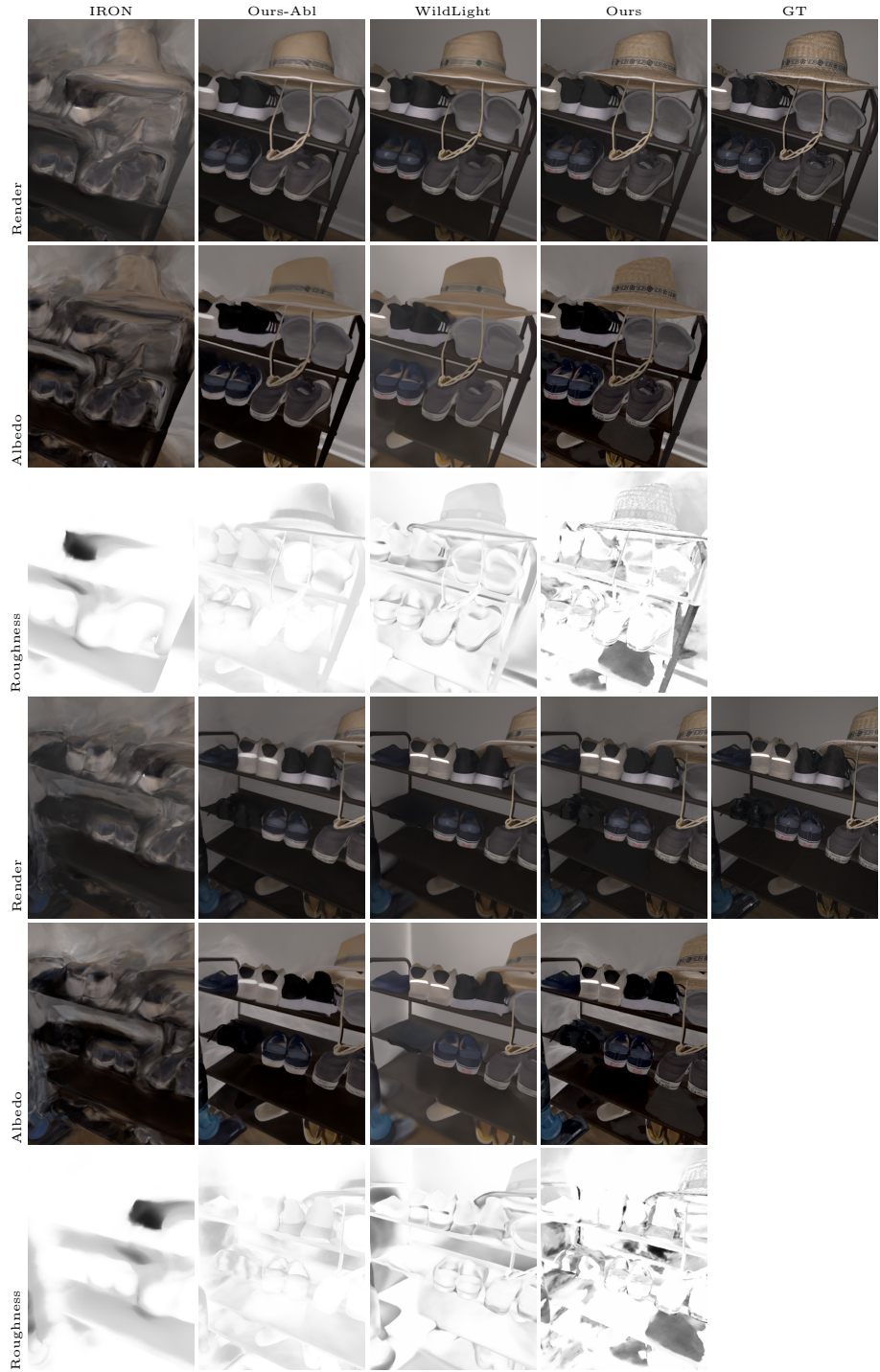}

\end{tabular}
\vspace{-10pt}
\caption{
Qualitative comparison on real scene shoe rack.
}
\label{fig:suppl_qualitative_figure_real_1}
\end{figure*}

\begin{figure*}
\footnotesize
\centering

\renewcommand{\tabcolsep}{1pt}
\begin{tabular}{cccccccc}

& IRON & Ours-Abl & WildLight & Ours & GT \\

\input{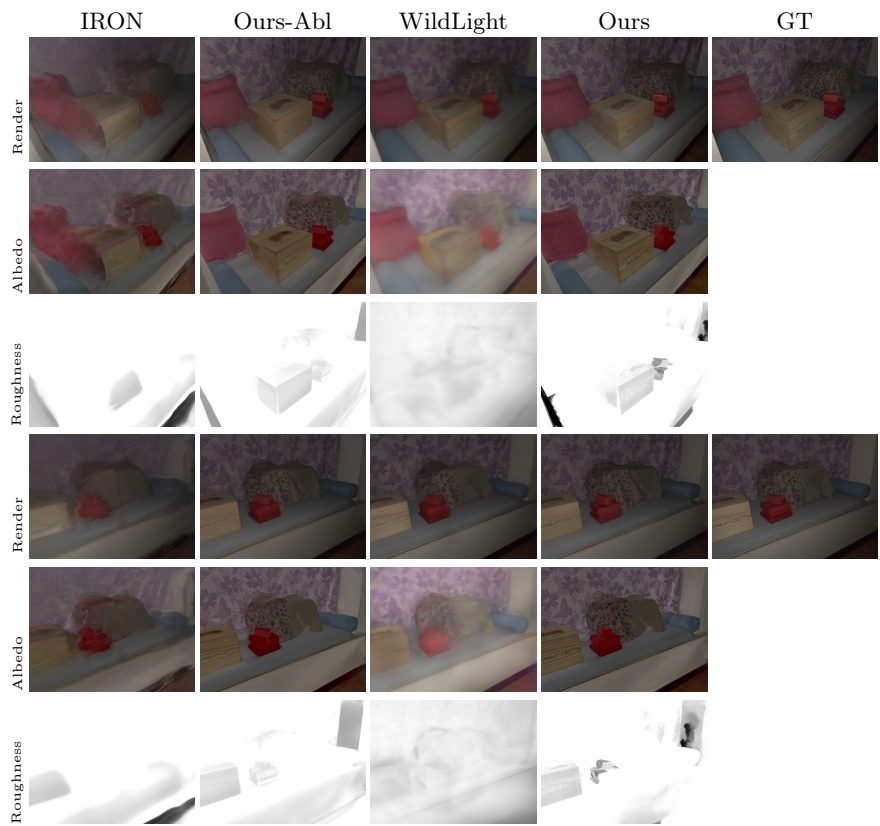}

\end{tabular}
\vspace{-10pt}
\caption{
Qualitative comparison on real scene window sill.
}
\label{fig:suppl_qualitative_figure_real_3}
\end{figure*}

In Figure~\ref{fig:suppl_qualitative_figure_synthetic_1}, Figure~\ref{fig:suppl_qualitative_figure_synthetic_2}, Figure~\ref{fig:suppl_qualitative_figure_synthetic_3}, we 
provide additional visualization of qualitative results on synthetic data. In Figure~\ref{fig:suppl_qualitative_figure_real_1}, Figure~\ref{fig:suppl_qualitative_figure_real_2}, Figure~\ref{fig:suppl_qualitative_figure_real_3}, Figure~\ref{fig:suppl_qualitative_figure_real_4}, we provide additional visualization of qualitative results on real data.

\providelength\width
\setlength\width{2.3cm}

\setlength\width{2.3cm}

\setlength\width{2.3cm}
\begin{figure*}
\footnotesize
\centering

\renewcommand{\tabcolsep}{1pt}
\begin{tabular}{cccccccc}

& IRON & Ours-Abl & WildLight & Ours & GT \\

\input{generated/suppl_qualitative_generated_real_1}

\end{tabular}
\vspace{-10pt}
\caption{
Qualitative comparison on real scene table.
}
\label{fig:suppl_qualitative_figure_real_2}
\end{figure*}

\setlength\width{2.2cm}

\setlength\width{2.2cm}
\begin{figure*}
\footnotesize
\centering

\renewcommand{\tabcolsep}{1pt}
\begin{tabular}{cccccccc}

& IRON & Ours-Abl & WildLight & Ours & GT \\

\input{generated/suppl_qualitative_generated_real_3}

\end{tabular}
\vspace{-10pt}
\caption{
Qualitative comparison on real scene coffee table.
}
\label{fig:suppl_qualitative_figure_real_4}
\end{figure*}

\clearpage

\end{document}